\documentclass[runningheads]{llncs}

\usepackage{eccv}

\usepackage{eccvabbrv}

\usepackage{graphicx}
\usepackage{booktabs}

\usepackage[accsupp]{axessibility}  

\usepackage{hyperref}

\usepackage{orcidlink}
\usepackage{float}
\usepackage{xcolor}
\usepackage[normalem]{ulem}
\usepackage{comment}
\usepackage{multirow}
\newcommand\XP[1]{\textcolor{black}{#1}}
\newcommand\AP[1]{\textcolor{black}{#1}}

\usepackage{appendix}

\begin{document}
\title{Multi-modal Crowd Counting via a Broker Modality} 

\titlerunning{Multi-modal Crowd Counting via a Broker Modality}

\author{Haoliang Meng\inst{1} \and
Xiaopeng Hong\inst{1,2}\thanks{Corresponding author.} \and
Chenhao Wang\inst{1} \and
Miao Shang\inst{1} \and
Wangmeng Zuo\inst{1,2}}

\authorrunning{H. Meng et al.}

\institute{Harbin Institute of Technology \and
Peng Cheng Laboratory\\
\email{menghaoliang2002@163.com}\  \email{hongxiaopeng@ieee.org} \ 
\email{22b903078@stu.hit.edu.cn}\\ \email{miaos0522@gmail.com}\quad \email{wmzuo@hit.edu.cn}}

\maketitle

\begin{abstract}

\XP{Multi-modal crowd counting involves estimating crowd density from both visual and thermal/depth images. This task is challenging due to the significant gap between these distinct modalities. In this paper, we propose a novel approach by introducing an auxiliary broker modality and on this basis frame the task as a triple-modal learning problem. We devise a fusion-based method to generate this broker modality, leveraging a non-diffusion, lightweight counterpart of modern denoising diffusion-based fusion models. Additionally, we identify and address the ghosting effect caused by direct cross-modal image fusion in multi-modal crowd counting. Through extensive experimental evaluations on popular multi-modal crowd counting datasets, we demonstrate the effectiveness of our method, which introduces only 4 million additional parameters, yet achieves promising results. The code is available at \url{https://github.com/HenryCilence/Broker-Modality-Crowd-Counting}.}
  
  \keywords{Multi-modal crowd counting \and Intermediate modality \and Modality fusion \and Diffusion model}
\end{abstract}

\section{Introduction}
\label{sec:intro}
Multi-modal crowd counting \cite{liu2021cross,peng2020rgb,lian2019density} is a challenging task which integrates information from various sources to provide a comprehensive understanding of complex crowds leveraging the complementary and commonality nature of diverse modalities.
However, there are two major challenges bringing difficulty for feature interaction and fusion. \textbf{On the one hand}, since vision and thermal images capture reflectance information in different bands of the spectrum, they look naturally different. Vision images excel at capturing color and texture information of visible light while thermal images are adept at containing information from invisible electromagnetic radiation. Therefore, vision images contain hardly any human figures under poor illumination conditions, while thermal images can clearly reflect the outline of each person under such conditions. Moreover, heating negative objects look almost the same in thermal images but can be easily distinguished in vision images. \textbf{On the other hand}, RGB and thermal images are not strictly aligned in position due to the difference in shooting angle. Directly fusing the two modalities causes a serious ghosting effect. The misalignment of RGB-thermal image pairs may lead to inaccuracy during the counting process, since ghosting may cause people to be counted twice. Therefore, simply extracting and fusing features of these two modalities during cross-modal learning may not perform well.

In order to solve these challenges, researchers have proceeded with abundant cross-modal feature interactions through the counting model. For example, Liu \etal propose a modality-shared branch to fully capture the shared information of different modalities. Pan \etal \cite{pan2024graph} feed thermal features to RGB image features to complete feature interaction. Liu \etal \cite{liu2023rgb} utilize a learnable count token to interact and fuse two-modal information at the scale level. However, all such methods directly perform feature interactions between the two original modalities and fail to achieve satisfactory performance. In contrast, we introduce an intermediate auxiliary modality to assist cross-modal learning. 

\begin{figure}[tb]
    \centering
    \includegraphics[width=0.8\textwidth]{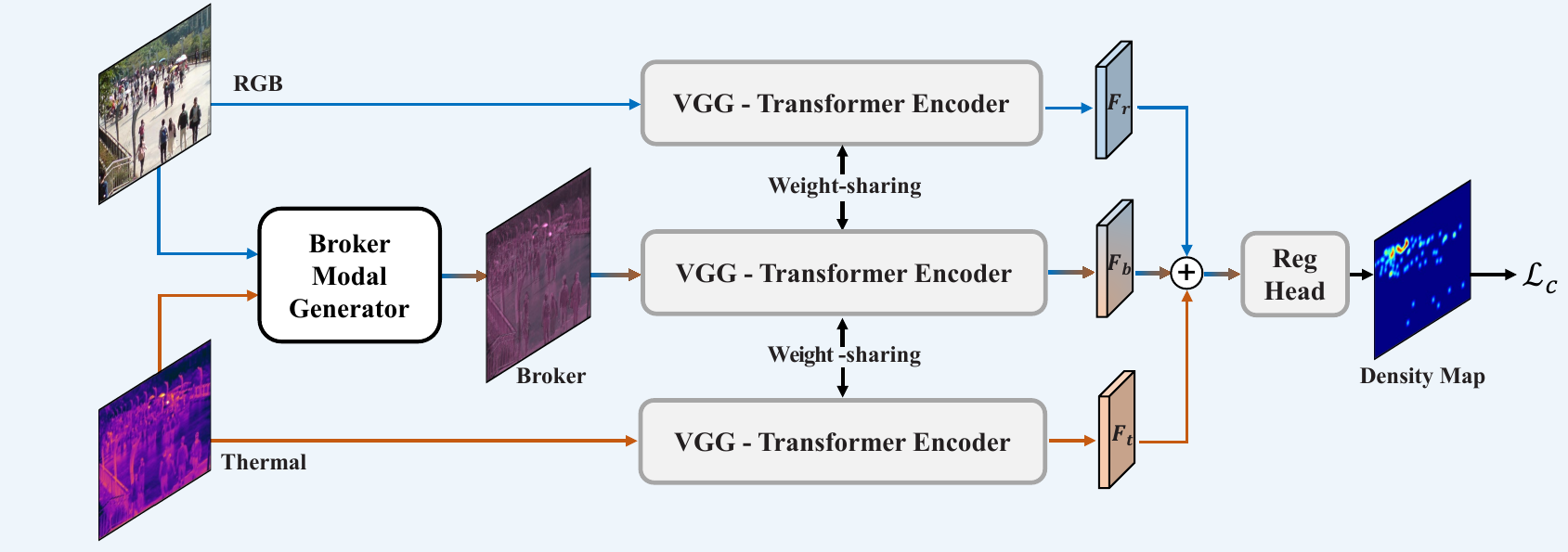}
    \caption{Illustration of the framework of our method. By leveraging the Broker Modal Generator (BMG) to introduce an auxiliary broker modality, we frame the dual-modal visual-thermal crowd counting task as a triple-modal learning problem.}
    \label{fig:1}
\end{figure}

\XP{Motivated by the analysis above, this paper presents a novel multi-modal crowd-counting approach. The basic idea is to transform a learning task from two modalities with a significant gap into a learning task involving three modalities by introducing an intermediate auxiliary modality, as shown in \cref{fig:1}. This intermediate modality, which we term a `broker' modal, serves to bridge and align the features of the two original modalities for smoother, more coherent, and effective fusion. While in the literature, methods for constructing intermediate modalities based on contrast learning have been used in similar tasks such as cross-modal person re-identification \cite{li2020infrared,zhang2021towards,wei2020co}, it is evident that contrast learning cannot be directly applied to regression tasks.}

\XP{To bridge this gap and generate the intermediate broker modality, we propose leveraging image fusion techniques, as the resulting fused image integrates information and characteristics from both modalities. The pivotal challenge lies in selecting the most appropriate fusion approach~\cite{zhao2023cddfuse,huang2022reconet,xu2023murf}. Among the available methods, recent denoising diffusion model-based fusion methods such as~\cite{zhao2023ddfm} stand out for their ability to preserve both structural and detailed information from the source images and achieve promising fusion results. However, these methods inherently entail a large-scale and computationally demanding nature, rendering them inefficient for direct integration as a module within our framework for generating intermediate modalities. To address this challenge, we introduce the idea of constructing a non-diffusion, lightweight counterpart of the denoising diffusion-based fusion model through a distillation-then-finetuning strategy. This model is initially built via a distillation process guided by denoising diffusion fusion models, followed by fine-tuning to tailor it to the crowd counting task.}
It inherits the high fusion capability of the denoising diffusion model while also addressing the low efficiency in training and fine-tuning associated with the diffusion model. Remarkably, we discover that such a lightweight non-diffusion model can accurately fuse the two modalities \XP{like a denoising diffusion model while being efficiently trained. Moreover, we identify the ghosting effect caused by directly fusing the two modalities and demonstrate that the designed distillation-then-finetuning learning scheme of our framework alleviates this phenomenon.}
We conduct comprehensive experiments on the popular multi-modal crowd counting benchmarks. Extensive experimental results demonstrate that our proposed approach significantly improves the counting performance.

The main contributions can be further summarized as follows:
\begin{itemize}
    \item \XP{We introduce an intermediate auxiliary modality and frame the dual-modal visual-thermal crowd counting task as a triple-modal learning problem.}
    \item We carefully design a broker modal generator, \XP{as a non-diffusion, lightweight counterpart of the denoising diffusion-based fusion model.}
    \item \XP{We provide an in-depth analysis of the ghosting effect caused by directly fusing the two modalities and provide a solution to alleviate this phenomenon.}
\end{itemize}

\section{Related Work}
\subsection{Multi-modal Crowd Counting}
With the development of deep learning technology, crowd counting has been increasingly studied and has become a significant computer vision task. There are three main technical routes among popular crowd counting methods: based on detection\cite{liu2018decidenet, sam2020locate, wang2021self, liu2023point}, based on regression\cite{liu2020semi, yang2020reverse} and based on density map\cite{ma2019bayesian,ma2020learning,lin2022boosting, lin2022semi, liu2018crowd, Liu_2019_ICCV}. Up to now, crowd counting methods for single modality have been well-studied. Researchers have made great achievements in ameliorating model structure\cite{lin2022boosting,lin2024gramformer,ma2020learning} and loss function\cite{ma2019bayesian,ma2021learning,lin2021direct}.

In recent years, with the development of multi-modal learning in various fields\cite{yu2022commercemm,chen2021topological,li2020comparison,ren2021learning}, multi-modal crowd counting\cite{pang2020hierarchical,wu2022multimodal,fan2020bbs,li2022rgb,zhang2020uc} has received widespread attention. Researchers use additional modalities such as thermal or depth together with visual images to enhance the performance of crowd counting methods. For example, Peng \etal \cite{peng2020rgb} propose a drone-based RGB-thermal multi-modal crowd counting dataset covering various attributes and design a multi-modal crowd counting network to learn cross-modal representation. Liu \etal \cite{liu2021cross} propose an RGB-thermal multi-modal crowd counting dataset containing complex environments and design a cross-modal collaborative representation learning framework to capture the complementary information of different modalities fully. Liu \etal \cite{liu2023rgb} use count-guided multi-modal fusion and modal-guided count enhancement to better excavate multi-modal information. Lian \etal \cite{lian2019density} collect a large-scale RGB-D crowd counting dataset and propose a regression-guided detection network for simultaneously crowd counting and localization. Li \etal \cite{li2022rgb} propose a cross-modal cyclic attention fusion model that integrates RGB and depth features in a cyclic attention manner to comprehensively model and fuse information from both modalities, which tackles the negative effect of the arbitrary crowd distribution on the counting task.

Most previous approaches focus on extracting features from two modalities and then interacting and fusing them to generate the density map. However, the significant gap between visible and infrared images brings difficulties to feature interaction and fusion. In this paper, we introduce an auxiliary intermediate modality to reduce the gap and participate in feature interaction and fusion.

\subsection{Intermediate Modality in Infrared-Visible Cross-Modal Tasks}
In various infrared-visible cross-modal tasks, one of the major challenges is the modality discrepancy between visible and infrared images. Aiming to mitigate the challenging modality discrepancy, quite a few recent approaches leverage intermediate spaces to bridge vision and infrared modalities. They design an intermediate modality generator to siphon off knowledge from visible and infrared images and output the intermediate images to assist the cross-modal process.

Li \etal \cite{li2020infrared} first propose an X modality generated by a lightweight network as an assistant to reduce modality discrepancy. Afterward, generators of various structures have been designed. For example, Zhang \etal \cite{zhang2021towards} propose a non-linear middle modality generator to project vision and infrared images into a unified middle modality image space and generate middle-modality images. Alehdaghi \etal \cite{alehdaghi2023adaptive} introduce an Adaptive Generation of Privileged Intermediate Information to adapt and generate a virtual domain that bridges discriminant information between vision and infrared modalities. Wang \etal \cite{wang2019learning} introduce an image-level sub-network to unify the representations for images with different modalities. Recently, Zhao \etal \cite{zhao2023metafusion} propose an infrared and visible image fusion method via meta-feature embedding for object detection, utilizing a fusion modality to assist the cross-modal detection process.

However, most approaches generate intermediate modality through contrastive learning and are designed for cross-modal person re-identification. Defining positive and negative pairs for contrastive learning in a dense regression task such as crowd counting is not straightforward. Consequently, applying such a scheme to cross-modal crowd counting is not feasible without significant modification. This presents an opportunity for us to introduce a new intermediate modality.

\section{Method}
\subsection{Counting Framework}

\AP{By introducing an auxiliary broker modality, we frame the dual-modal visual-thermal crowd counting task as a triple-modal learning problem.}
As illustrated in \cref{fig:1}, our counting model consists of three parts: a broker modal generator (BMG), a feature extractor, and a regression head. The feature extractor is implemented by an architecture of ‘VGG-19 - Transformer encoder’ and is weight-sharing for all modalities. We first use BMG to generate the broker modality $F\in \mathbb{R}^{3\times H\times W}$, then feed it to the feature extractor along with two source modalities $R$ and $T\in \mathbb{R}^{3\times H\times W}$. 
\begin{equation}
  F = g\left(R,T\right),
\end{equation}

\begin{equation}
  F_{r} = \zeta_{v}\left(\zeta_{c}\left(R\right)\right),
  F_{f} = \zeta_{v}\left(\zeta_{c}\left(F\right)\right),
  F_{b} = \zeta_{v}\left(\zeta_{c}\left(T\right)\right),
\end{equation}

where $g$ denotes the broker modal generator, $R$ and $T$ denote RGB and thermal images, and $F_{r}, F_{f}, F_{b}\in \mathbb{R}^{D\times h\times w}$ are features from three modalities. $\zeta_{c}$ and $\zeta_{v}$ denote VGG-19 and Transformer encoder in the feature extractor, respectively.

Finally, we \XP{sum the features up and send the summation} to the regression head $R$ to obtain a one-channel density map $D^{est}\in \mathbb{R}^{1\times h\times w}$.

Considering the performance and robustness, we adopt the popular Bayesian Loss \cite{ma2019bayesian} as the loss function to align discrete point annotations and the continuous density map. 
\begin{equation}
  \mathcal{L}_{c} = \sum_{i=1}^{M}|1-\langle\frac{\mathcal{N}\left(z_{i},\sigma^{2}\mathbf{1}_{2\times2}\right)}{\sum\limits_{n=1}^{M}\mathcal{N}\left(z_{n},\sigma^{2}\mathbf{1}_{2\times2}\right)},D^{est}\rangle|,
\end{equation}
where $M$ is the total crowd count of the image and $z_{n}$ is the $n$-th annotated head point position in the image. $\langle\cdot,\cdot\rangle$ denotes the inner product operation.

\subsection{Broker Modality Generator}

\AP{We introduce an auxiliary broker modal to bridge RGB and thermal modalities. Unlike most contrastive learning-based methods mentioned above \cite{li2020infrared,zhang2021towards}, our auxiliary modal is constructed on multi-modal image fusion. We choose DDFM \cite{zhao2023ddfm}, one state-of-the-art multi-modal image fusion method as the reference to construct our Broker Modal Generator (BMG). Nonetheless, the denoising diffusion model usually suffers from high computation costs in both training and prediction, making it inefficient to serve as a part of an end-to-end counting mode. Moreover, it directly fuses original images without registration, causing the ghosting effect in fusion images. To alleviate this issue in an end-to-end training and prediction multi-modal counting framework, we design a lightweight non-diffusion generator BMG as its counterpart and suit it for the task. We design a two-stage learning scheme for BMG, which consists of a distillation stage and a fine-tuning stage.  In the following sections, we will elaborate on the structure of the broker modality generator and its learning scheme.}

\begin{figure}[tb]
    \centering
    \includegraphics[width=0.8\textwidth]{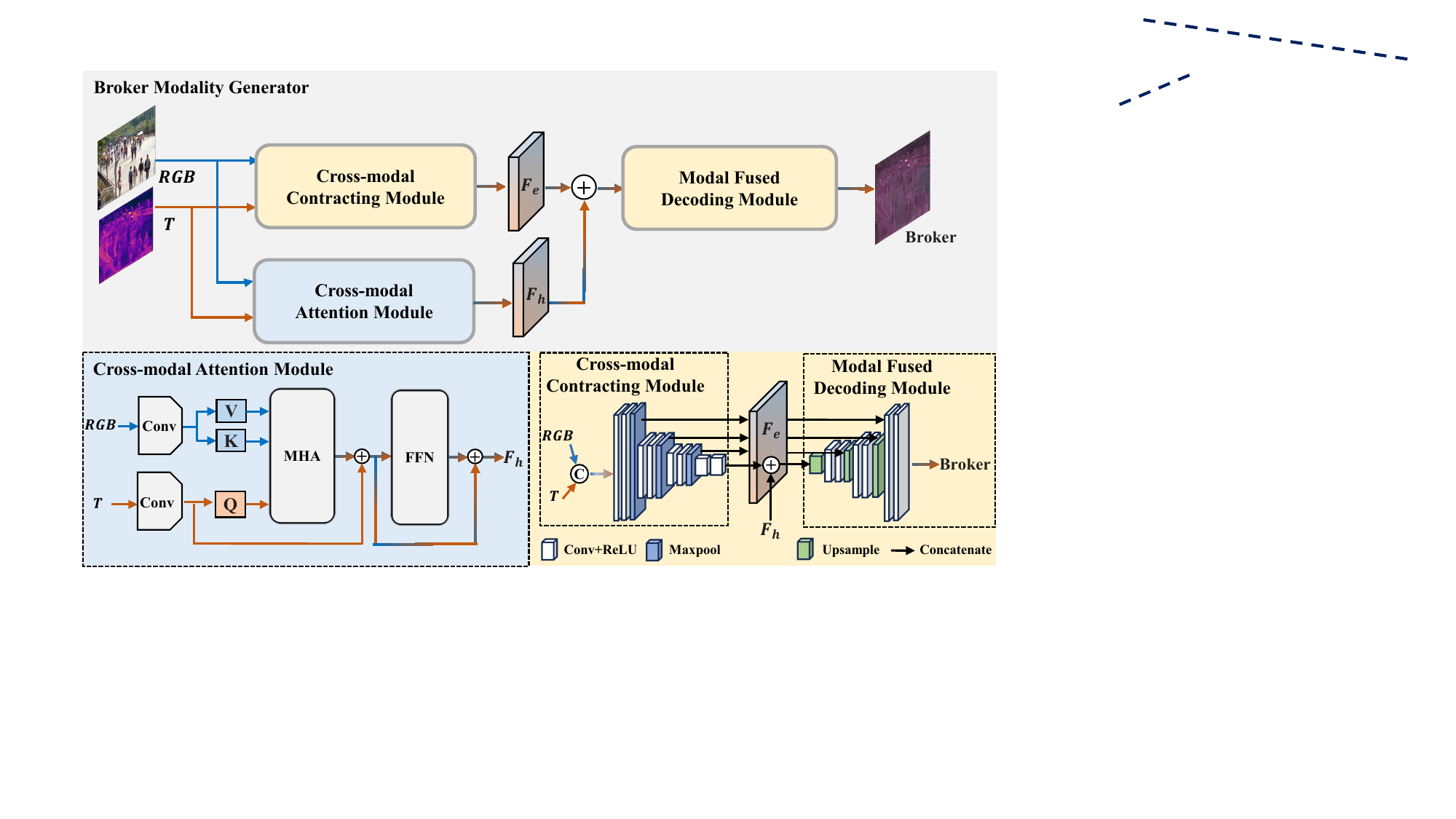}
    \caption{The framework of our broker modal generator. Specifically, the generator consists of a cross-modal contracting module and a modal fused decoding module for image reconstruction, and a cross-modal attention module to enhance modality alignment.} 
    \label{fig:2}
\end{figure}

\noindent \textbf{The structure of BMG.}
As illustrated in \cref{fig:2}, the generator consists of three parts: A cross-modal contracting module (CMC), a modal fused decoding module (MFD), and a cross-modal attention module (CMA). The cross-modal contracting module $f_{e}$ and modal fused decoding module $f_{d}$ are implemented by U-Net\cite{10.1007/978-3-319-24574-4_28} encoder and decoder, respectively, while the cross-modal attention module is implemented by a 1-layer, 4-head cross-attention Transformer encoder. Firstly, we use two $3\times3$ convolutional layers $\xi_{r}$ and $\xi_{t}$ to extract the feature of RGB and thermal images. Then, we concatenate them together as the input of CMC $f_{e}$ and obtain the encoding result $F_{e}\in\mathbb{R}^{d\times h\times w}$:
\begin{equation}
  F_{e}=f_{e}\left( [\xi_{r}\left(R\right); \xi_{t}\left(T\right)]\right),
\end{equation}
where $[ \ \cdot \ ; \ \cdot \ ]$ denotes the channel concatenating operation.

Furthermore, we use cross-attention to enhance the encoding results of CMC. We use convolutional layers to extract feature maps of two original modalities and flatten them to a sequence of patch embedding $X_{r}$, $X_{t}\in\mathbb{R}^{hw\times d}$. We use T modal embedding as Query(Q) and RGB modal embedding as Key(K) and Value(V) for the multi-head cross-attention operation. Note that the residual connection in cross-attention is contributed by the T modal. 

\begin{equation}
\begin{aligned}
  X_{r}&=\xi\left(R\right), \ X_{t}=\xi\left(T\right),\\
  Q&=X_{t}W_{q}, \
  K=X_{r}W_{k}, \
  V=X_{r}W_{v},\\
  H&=\textit{Att}\left(Q,K,V\right) + X_{t},\\
  F_{h}&=\textit{FFN}(H)+H,\\
\end{aligned}
\end{equation}

\noindent where $\xi$ is the convolutional layer in the cross-modal attention module, $W_{q}$,$W_{k}$,$W_{v}$ are the projection weights for multi-head attention layer, FFN denotes the feedforward network in Transformer encoder, and
\begin{equation}
  \textit{Att}\left(Q,K,V\right)= \textit{Softmax}\left(\frac{QK^T}{\sqrt{d}}\right)V.
\end{equation}

Finally, we add the enhancement result $F_{h}\in\mathbb{R}^{d\times h\times w}$ to the encoding result $F_{e}$ as the input of MFD and obtain the multi-modal fusion results $F\in\mathbb{R}^{3\times H\times W}$:

\begin{equation}
  F=f_{d}\left(F_{e}+F_{h}\right).
\end{equation}

\noindent \textbf{The distillation-then-finetuning learning scheme of BMG.}

\begin{figure}[tb]
    \centering
    \includegraphics[width=0.8\textwidth]{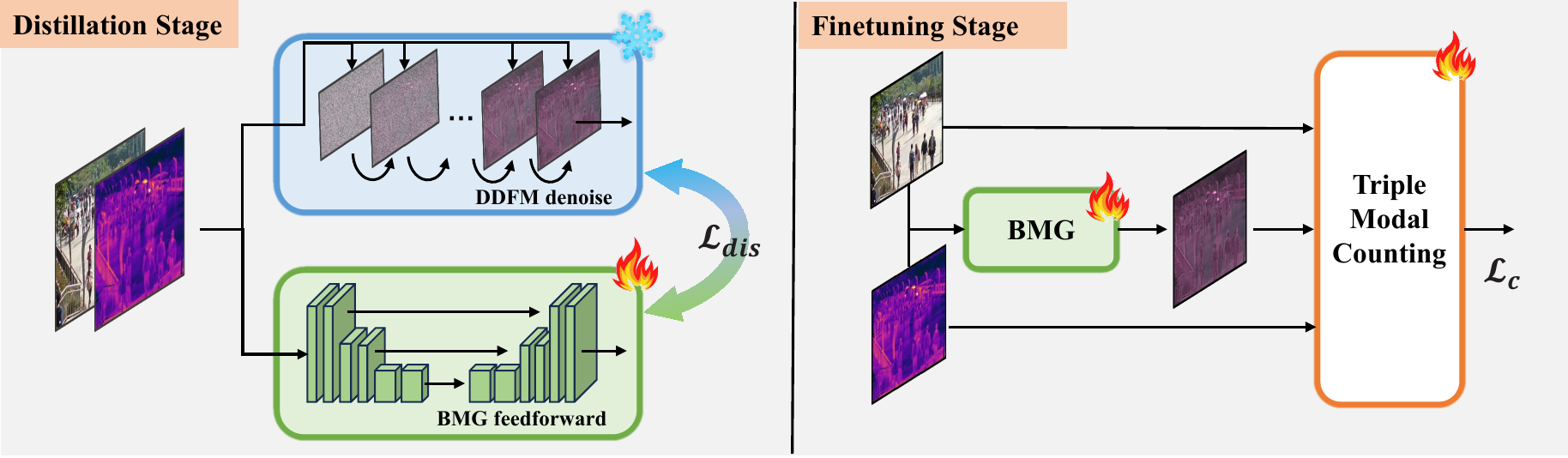}
    \caption{Illustration of the distillation-then-finetuning strategy. In the distillation stage, BMG is initialized using a distillation process guided by DDFM. In the fine-tuning stage, BMG is tuned together with the feature extractor to suit the counting task.}
    \label{fig:d}
\end{figure}

The learning process consists of a distillation stage and a fine-tuning stage. During the distillation stage, we pre-train BMG by distilling DDFM as its counterpart. During the fine-tuning stage, we tune BMG together with the feature extractor to better suit the counting task and bridge two original modalities.

\AP{\it{The distillation stage.}}
\AP{Before counting, we first pre-train BMG using MSE Loss in the first stage to enable it to acquire the infrared-visible fusion capability, as shown in \cref{fig:d}. We achieve this through distilling DDFM\footnote{For the infrared-visible fusion task, the DDFM model retains both structural and detailed information from the source images, meeting the visual fidelity requirements as well. Thus it is used to pre-train our broker modal generator based on its favorable multi-modal fusion results.} \cite{zhao2023ddfm}, one state-of-the-art image fusion model. During the distillation stage, BMG \XP{approximates} the high fusion capability of DDFM with a much lower scale and efficiency.}

\begin{equation}
  \mathcal{L}_{dis} = \sum_{i=1}^{K}\left|\left|g\left(R,T\right)- p\left(R,T\right)\right|\right|_{2}^{2},
\end{equation}
where $p$ denotes DDFM, and $K$ is the number of training samples.

\AP{\it{The fine-tuning stage.}}
\AP{In the fine-tuning stage, we generate the broker modal by BMG and send it to the feature extractor together with the original modalities, framing the dual-modal counting task as a triple-modal learning problem. We tune BMG together with the \XP{feature extractor} during the counting process to suit the counting task. During fine-tuning, BMG acquires the capability to bridge the two original modalities coherently and to alleviate the ghosting phenomenon caused by space misalignment.} The influence and effect of the two-stage learning process are shown in \XP{Section} 4.5.

\subsection{Discussion on the ghosting effect.}

\begin{figure}[tb]
    \centering
    \includegraphics[width=0.8\textwidth]{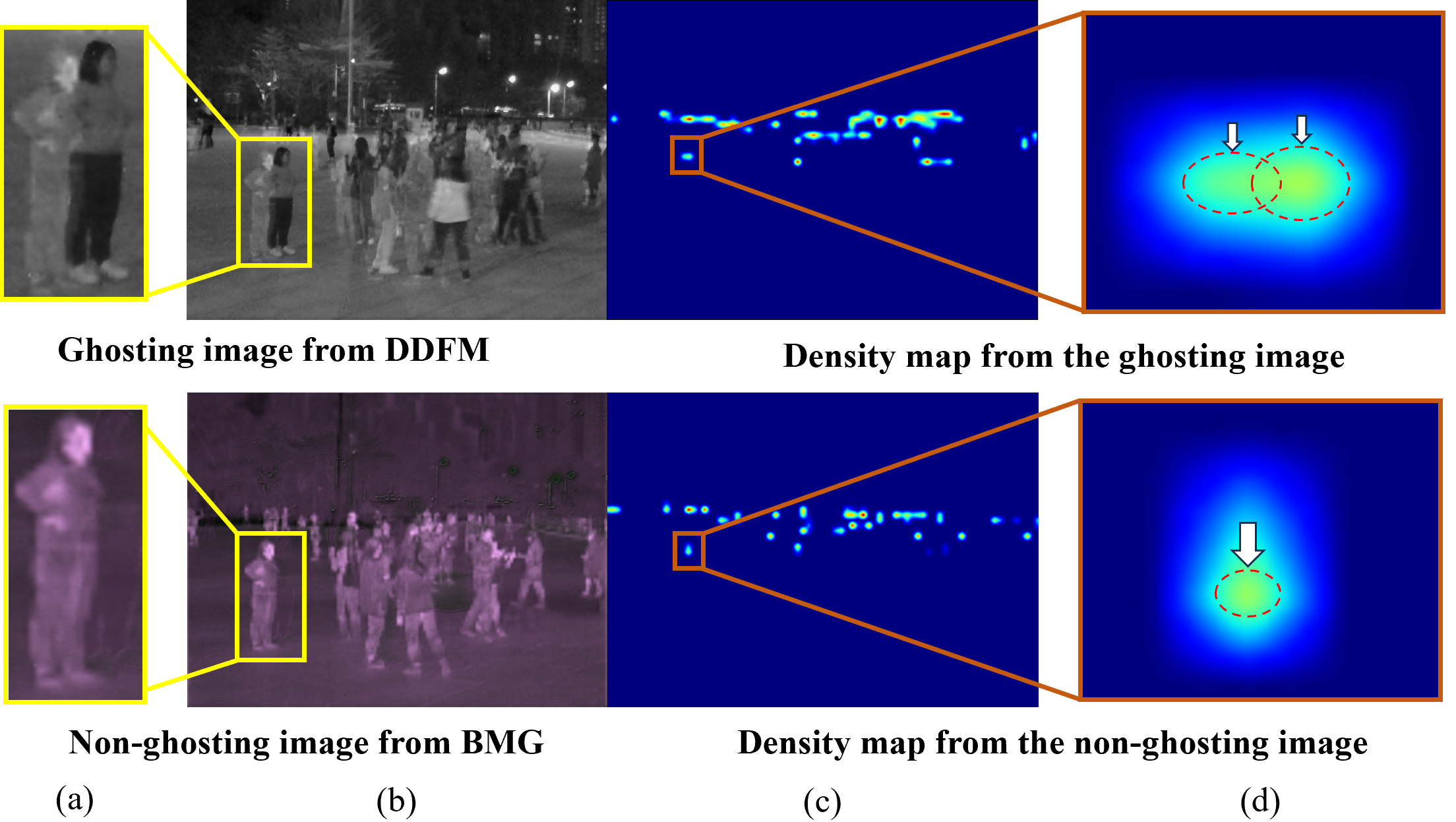}
    \caption{Comparison of ghosting and non-ghosting intermediate images and the corresponding density maps. (b) are the intermediate images generated by DDFM directly and by our BMG, and (a) shows the comparison of the ghosting region in (b). (c) are the corresponding density maps generated from the scenes in (b), and (d) shows the difference of the density maps in the ghosting region. } 
    \label{fig:7}
\end{figure}

\XP{As visual and thermal images are captured by two distinct sensors, there exists an angular disparity or spatial displacement between them. Therefore, fusing them directly using DDFM may lead to a ghosting effect\footnote{To better illustrate the challenges posed by this problem, we conducted experiments to assess the effectiveness of image alignment-based de-ghosting algorithms such as \cite{ma2015robust,gao2017image,jiang2020contour}. Our results indicate that while satisfactory outcomes can be attained with natural image pairs, the de-ghosting algorithm performs inadequately on image pairs with low imaging quality, such as the data used in this study. Further comprehensive experiments are available in the supplementary material.} in the fusion image. For instance, in the first line of \cref{fig:7}(a), two overlapping figures for one person are evident in the yellow rectangular area. In such cases, counting models may mistakenly interpret such a fusion region as a region of two separate entities, resulting in a density region containing overlapping Gaussian distributions, as depicted in the first line of \cref{fig:7}(d).}

\noindent \textbf{Ghosting effect alleviation by two-stage learning.}
\XP{The proposed two-stage learning scheme, somehow surprisingly, mitigates this issue well. The key is to decouple the distillation of denoising diffusion models from the counting process. In the distillation stage, the training process exclusively distills a heavy denoising diffusion multi-modal image fusion model (with a size of over 500M, as shown in \cref{tab:5}) using a lightweight network (with a size of less than 4M). This single training task ensures the convergence of the training process. In the fine-tuning stage, the training process fine-tunes the pre-trained broker modal generator to the counting task, facilitating the elimination of the ghosting effect. Specifically, as the analysis above suggests, the ghosting effect may cause the model to generate overlapping Gaussian distributions and overestimate the count. By calculating the counting loss, which directly reflects the produced density map's deviation from the ground truth, the error caused by the ghosting effect is penalized during fine-tuning. Consequently, this leads to a clearer fusion image with a significantly reduced ghosting effect.}

\section{Experiments and Discussion}
\label{sec:blind}
\subsection{Evaluation Metrics and Datasets}

We conduct our experiments on three widely used multi-modal crowd counting benchmark datasets: RGBT-CC, DroneRGBT, and ShanghaiTech RGBD. We adopt RMSE\cite{sindagi2019multi} and GAME\cite{guerrero2015extremely} to evaluate the methods.

\noindent \textbf{RGBT-CC} \cite{liu2021cross} is the most widely used large-scale RGBT crowd counting benchmark, in which 138,389 pedestrians are annotated in 2,030 pairs of RGB-thermal images captured in unconstrained scenarios. All image pairs are captured by an optical-thermal camera in various scenarios like malls, streets, playgrounds, train stations, and metro stations. 

\noindent \textbf{DroneRGBT} \cite{peng2020rgb} is a drone-based RGBT crowd counting benchmark dataset. It contains 3,607 registered RGB-thermal image pairs with ground truth annotations and covers different attributes including height, illumination, and density. Images are captured by drone-mounted cameras, covering a wide range of scenarios like campus, street, park, parking lot, playground, and plaza.

\noindent \textbf{ShanghaiTechRGBD} \cite{lian2019density} is a large-scale RGBD crowd counting dataset containing 2,193 RGB-depth image pairs with 144,512 annotations. Images are captured by a stereo camera whose valid depth ranges from 0 to 20 meters. The scenes include busy streets of metropolitan areas and crowded public parks.

\subsection{Implementation Details}
\noindent \textbf{Network structure.}
CMC of BMG downsamples the input image size $(H,W)$ to $(h,w)$ while increasing channels from 6 to $d=256$. The decoder upsamples the features to the full input size $(H,W)$ and reconstructs a 3-channel broker modal $F$. The patch number for cross-modal attention is 64 and embedding dimension $d=256$. The number of the Transformer encoder layer is 2 and the head number is 6. The patch number of the Transformer encoder is 196 and the embedding dimension is $D=768$. We upsample the output of the feature extractor to $(h,w)$, and then feed it to the regression head.

\noindent \textbf{Experimental setup and training details.}
We implement our model with PyTorch and train it on an NVIDIA RTX 3090 GPU with 24GB memory. We adopt the Adam optimizer with the learning rate of 1e-5 and the weight decay of 1e-4. During training, we randomly crop and flip the input images for data augmentation. The crop size is $224\times224$ for RGBT-CC and DroneRGBT, and $1024\times1024$ for ShanghaiTechRGBD. The batch size for all datasets is 1 and the maximum number of training epochs is 400.

\begin{table}[tb]
\scriptsize
  \caption{Comparison with the state-of-the-art methods on RGBT-CC.
  }
  \label{tab:1}
  \centering
  \begin{tabular}{ccccccc}
    \toprule
    Method & Venue & GAME(0) & GAME(1) & GAME(2) & GAME(3) & RMSE\\
    \midrule
    UCNet\cite{zhang2020uc} & CVPR 2020 & 33.96 & 42.42 & 53.06 & 65.07 & 56.31\\
    HDFNet\cite{pang2020hierarchical} & ECCV 2020 & 22.36 & 27.79 & 33.68 & 42.48 & 33.93\\   
    MVMS\cite{zhang2019wide} & CVPR 2019 & 19.97 & 25.10 & 31.02 & 38.91 & 33.97\\
    BBSNet\cite{fan2020bbs} & ECCV 2020 & 19.56 & 25.07 & 31.25 & 39.24 & 32.48\\
    CmCaF\cite{li2022rgb} & TII 2022 & 15.87 & 19.92 & 24.65 & 28.01 & 29.31\\
    BL+IADM\cite{liu2021cross} & CVPR 2021 & 15.61 & 19.95 & 24.69 & 32.89 & 28.18\\
    CSCA\cite{zhang2022spatio} & ACCV 2022 & 14.32 & 18.91 & 23.81 & 32.47 & 26.01\\
    BL+MAT\cite{wu2022multimodal} & ICME 2022 & 13.61 & 18.08 & 22.79 & 31.35 & 24.48\\
    TAFNet\cite{tang2022tafnet} & ISCAS 2022 & 12.38 & 16.98 & 21.86 & 30.19 & 22.45\\
    BL+MAT+SSP\cite{wu2022multimodal} & ICME 2022 & 12.35 & 16.29 & 20.81 & 29.09 & 22.53\\
    GETANet\cite{pan2024graph} & GRSL 2024 & 12.14 & 15.98 & 19.40 & 28.61 & 22.17\\
    DEFNet\cite{zhou2022defnet} & TITS 2022 & 11.90 & 16.08 & 20.19 & 27.27 & 21.09\\
    R2T\cite{li2022learning} & KBS 2022 & 11.63 & 16.70 & 22.12 & 32.32 & 21.28\\
    MJPNet-T\cite{zhou2024mjpnet} & IoT 2024 & 11.56 & 16.36 & 20.95 & 28.91 & 17.83\\
    MCN\cite{kong2024cross} & ESWA 2024 & 11.56 & 15.92 & 20.16 & 28.06 & 19.02\\
    $\mathrm{MC^3Net}$\cite{zhou2023mc} & TITS 2023 & 11.47 & 15.06 & 19.40 & 27.95 & 20.59\\  
    C4-MIM\cite{guo2024consistency} & CIS 2024 & 11.08 & 14.85 & 19.05 & 24.94 & 19.71\\
    CAGNet\cite{yang2024cagnet} & ESWA 2024 & 11.06 & 14.73 & 18.94 & 25.76 & 17.83\\
    BGDFNet\cite{xie2024bgdfnet} & TIM 2024 & 11.00 & 15.04 & 19.86 & 29.72 & 19.05\\
    VPMFNet\cite{mu2024visual} & IoT 2024 & 10.99 & 15.17 & 20.07 & 28.03 & 19.61\\
    MSDTrans\cite{liu2023rgb} & BMVC 2022 & 10.90 & 14.81 & 19.02 & 26.14 & 18.79\\
    \midrule
    Ours & ECCV 2024 & {\bf 10.19} & {\bf13.61} & {\bf17.65} & {\bf23.64} & {\bf17.32}\\
  \bottomrule
  \end{tabular}
\end{table}

\begin{table}[tb]
\scriptsize
  \caption{Comparison with the state-of-the-art methods on DroneRGBT.
  }
  \label{tab:2}
  \centering
  \begin{tabular}{cccc}
    \toprule
    Method & Venue & GAME(0) & RMSE\\
    \midrule
    CMDBIT\cite{xie2023cross} & TCSVT 2023 & 11.50 & 21.27\\
    BL+IADM\cite{liu2021cross} & CVPR 2021 & 11.41 & 17.54\\
    DEFNet\cite{zhou2022defnet} & TITS 2022 & 10.87 & 17.93\\
    CSRNet\cite{li2018dilated} & CVPR 2018 & 8.91 & 13.80\\
    GETANet\cite{pan2024graph} & GRSL 2024 & 8.44 & 13.99\\
    CGINet\cite{pan2023cginet} & EAAI 2023 & 8.37 & 13.45\\
    $\mathrm{MC^3Net}$\cite{zhou2023mc} & TITS 2023 & 7.33 & 11.17\\
    MMCCN\cite{peng2020rgb} & ACCV 2020 & 7.27 & 11.45\\
    I-MMCCN\cite{zhang2021mmccn} & IC-NIDC 2021 & 6.91 & 11.26\\
    \midrule
    Ours & ECCV 2024 & {\bf 6.20} & {\bf10.40}\\
  \bottomrule
  \end{tabular}
\end{table}

\subsection{Experimental Evaluations}

\begin{table}[tb]
\scriptsize
  \caption{Comparison with the state-of-the-art methods on ShanghaiTechRGBD.
  }
  \label{tab:3}
  \centering
  \begin{tabular}{ccccccc}
    \toprule
    Method & Venue & GAME(0) & GAME(1) & GAME(2) & GAME(3) & RMSE\\
    \midrule
    UCNet\cite{zhang2020uc} & CVPR 2020 & 10.81 & 15.24 & 22.04 & 32.98 & 15.70\\
    DetNet\cite{liu2018decidenet} & CVPR 2018 & 9.74 & - & - & - & 13.14\\    
    HDFNet\cite{pang2020hierarchical} & ECCV 2020 & 8.32 & 13.93 & 17.97 & 22.62 & 13.01\\   
    CL\cite{idrees2018composition} & ECCV 2018 & 7.32 & - & - & - & 10.48\\  
    GETANet\cite{pan2024graph} & GRSL 2024 & 6.91 & 8.83 & 12.31 & 17.11 & 9.88\\
    BL+MAT\cite{wu2022multimodal} & ICME 2022 & 6.61 & 8.16 & 10.09 & 16.62 & 9.39\\
    BBSNet\cite{fan2020bbs} & ECCV 2020 & 6.26 & 8.53 & 11.80 & 16.46 & 9.26\\
    BL+MAT+SSP\cite{wu2022multimodal} & ICME 2022 & 5.39 & 6.73 & 8.98 & 13.66 & 7.77\\
    VPMFNet\cite{mu2024visual} & IoT 2024 & 5.24 & 6.85 & 9.14 & 13.81 & 8.19\\
    RDNet\cite{lian2019density} & CVPR 2019 & 4.96 & - & - & - & 7.22\\
    CSCA\cite{zhang2022spatio} & ACCV 2022 & 4.39 & 6.47 & 8.82 & 11.76 & 6.39\\
    CSRNet+IADM\cite{liu2021cross} & CVPR 2021 & 4.38 & 5.95 & 8.02 & 11.02 & 7.06\\
    DPDNet\cite{lian2021locating} & TPAMI 2021 & 4.23 & 5.67 & 7.04 & 9.64 & 6.75\\
    AGCCM\cite{mo2022attention} & TIP 2022 & 3.90 & - & - & - & 5.86\\
    CCANet\cite{liu2023ccanet} & TMM 2023 & 3.78 & 5.34 & 7.63 & 10.17 & 5.56\\
    \midrule
    Ours & ECCV 2024 & \bf{3.72} & {6.06} & {9.02} & {12.63} & {\bf5.28}\\
  \bottomrule
  \end{tabular}
\end{table}

\begin{table}[thbp!]
\scriptsize
  \caption{Comparison of parameter quantity with the state-of-the-art models}
  \label{tab:6}
  \centering
  \begin{tabular}{cccc}
    \toprule
    Method & Venue & Param(M)$\downarrow$\\
    \midrule
    HDFNet\cite{pang2020hierarchical} & ECCV 2020 & 45.63\\
    UCNet\cite{zhang2020uc} & CVPR 2020 & 31.08\\
    BBSNet\cite{fan2020bbs} & ECCV 2020 & 49.78\\
    BL+IADM\cite{liu2021cross} & CVPR 2021 & 25.67\\
    DEFNet\cite{zhou2022defnet} & TITS 2022 & 45.33\\
    $\mathrm{MC^3Net}$\cite{zhou2023mc} & TITS 2023 & 113.08\\   
    \midrule
    Ours & ECCV 2024 & 40.55\\
  \bottomrule
  \end{tabular}
\end{table}

We compare our proposed method with the state-of-the-art methods on the benchmark datasets described above. 

\begin{figure}[thbp!]
    \centering
    \includegraphics[width=0.8\textwidth]{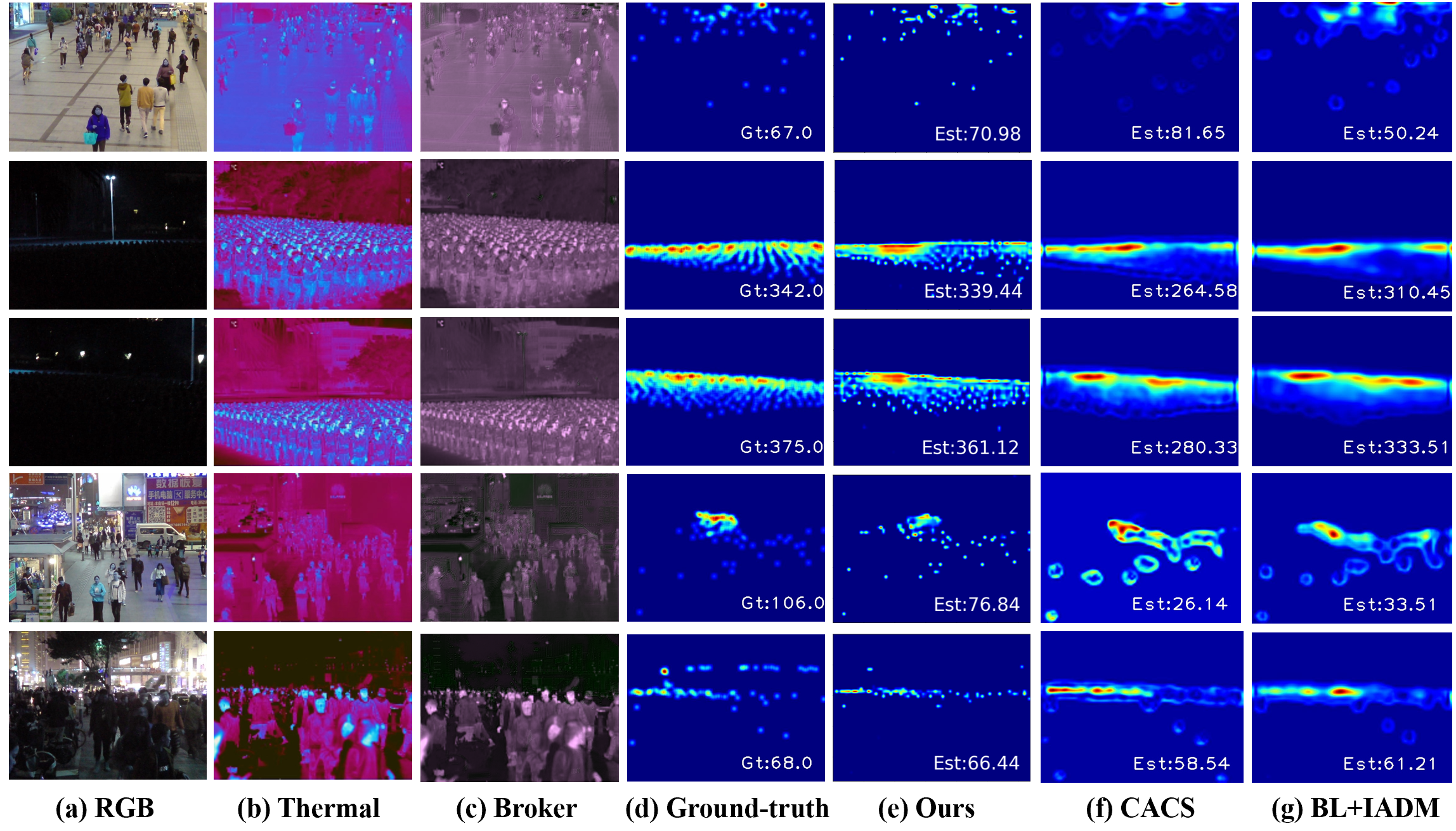}
    \caption{Visualization results of our method compared with different models. Our method consistently outperforms compared with other previous state-of-the-art models, demonstrating the effectiveness and superiority of the proposed model in accurately estimating multi-modal crowd density.}
    \label{fig:3}
\end{figure}

\noindent \textbf{Experimental results on RGB-T datasets.}
Experimental results on RGBT-CC and DroneRGBT are shown in \cref{tab:1} and \cref{tab:2}, respectively. The experimental results demonstrate that our method \XP{clearly outperforms} previous state-of-the-art models and becomes the new state-of-the-art method on both RGB-T datasets. For example, compared to MSDTrans\cite{liu2023rgb}, our method shows significant improvement on all five evaluation metrics on RGBT-CC. It reduces GAME(0), GAME(1), GAME(2), GAME(3), and RMSE by 0.71, 1.20, 1.37, 2.50, and 1.47, respectively. Compared to I-MMCCN \cite{zhang2021mmccn} on DroneRGBT, it improves achievement of 0.71 and 0.86 on GAME(0) and RMSE, respectively. Furthermore, we provide visual comparison results with previous state-of-the-art methods in \cref{fig:3}. These quantitative and qualitative experiments demonstrate that our proposed method is greatly effective and robust.

\noindent \textbf{Experimental results on RGB-D dataset.}
Furthermore, to thoroughly evaluate the effectiveness and robustness of our proposed multi-modal crowd counting approach, we conducted experiments in the realm of RGB-depth crowd counting tasks. Experimental results on ShanghaiTechRGBD are shown in \cref{tab:3}. Notably, our approach achieves the best GAME(0) value of 3.72 and the best RMSE value of 5.28. These compelling results underscore the robustness and versatility of our proposed method in handling the intricacies associated with multi-modal crowd counting tasks. 

\noindent \textbf{Parameter quantity of different models.}
We also analyze the complexity of models by comparing the parameter quantity with state-of-the-art methods, as in \cref{tab:6}. 
Despite achieving remarkable results in terms of crowd counting accuracy, it is noteworthy that our model maintains an acceptable and competitive parameter quantity when benchmarked against contemporary state-of-the-art methodologies.

In general, the results above demonstrate that our method achieves high accuracy in both RGB-T and RGB-D crowd counting tasks. These results show the remarkable robustness and effectiveness of our approach in addressing the challenging multi-modal crowd counting task.

\begin{table}[tb]
\scriptsize
  \caption{Influence of the distillation-then-finetuning training scheme for BMG.}
  \label{tab:fine-tune}
  \centering
  \begin{tabular}{c|c|ccccc}
    \toprule
    Distillation stage& Fine-tuning stage & GAME(0) & GAME(1) & GAME(2) & GAME(3) & RMSE\\
    \midrule
    $\times$ & \checkmark & 12.93 & 16.12 & 20.27 & 26.07 & 25.49\\
    \checkmark & $\times$ & 11.29 & 14.52 & 18.70 & 24.41 & 20.85\\
    \checkmark & \checkmark& {\bf 10.19} & {\bf13.61} & {\bf17.65} & {\bf23.64} & {\bf17.32}\\
  \bottomrule
  \end{tabular}
\end{table}

\subsection{Influence of Distillation-then-finetuning Training Scheme}

\begin{figure}[th]
    \centering
    \includegraphics[width=0.8\textwidth]{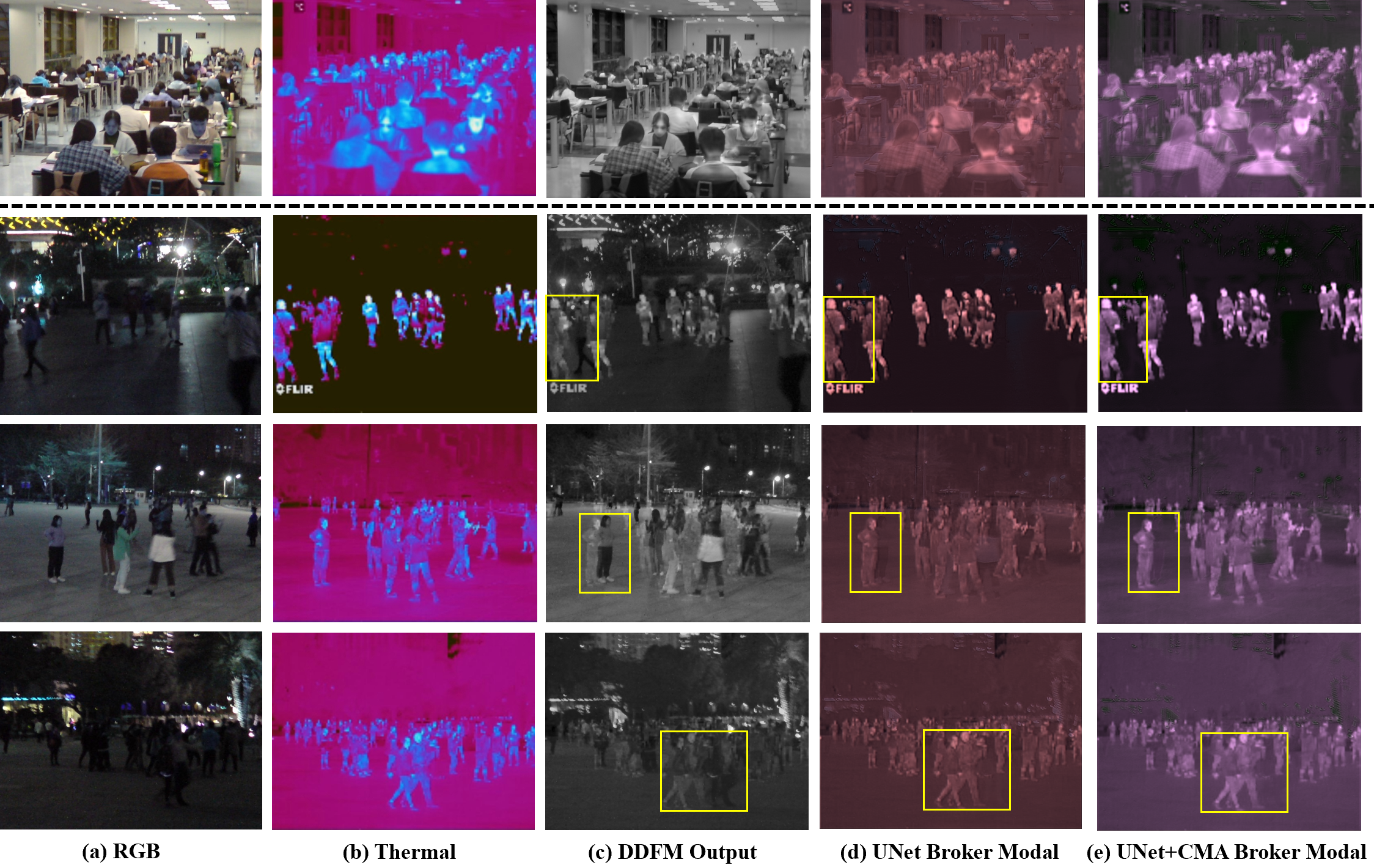}
    \caption{Original and broker modalities generated by different generators. The first row is the case where the space misalignment is slight, and the remaining is the case where the space misalignment is relatively significant. 
    The broker modal harmoniously bridges the attributes of the two source modalities and mitigates the gap between them. Moreover, the ghosting caused by misalignment in space is greatly eliminated during the fine-tuning stage, as shown in the yellow rectangular area.}
    \label{fig:f}
\end{figure}

Here we investigate the influence of designing the distillation-then-finetuning training scheme for BMG. First, \cref{tab:fine-tune} shows that the two-stage training scheme greatly improves the performance of the model. If training without DDFM distillation initialization or fine-tuning BMG, the performance obviously decreases (\emph{e.g.} GAME(0) from 10.19 to 11.29 and 12.93). This validates both the distillation and fine-tuning stages in two-stage learning. Besides, \cref{tab:5} shows the performance of counting using different intermediate modalities, the time cost of generating one intermediate image, and the parameter quantity of different generators. Compared with DDFM, our BMG achieves superior results with a more economical cost. Moreover, the simple and elegant fine-tuning of BMG unexpectedly alleviates the ghosting phenomenon, as shown in \cref{fig:f}

\begin{figure}[tb]
    \centering
    \includegraphics[width=1.0\textwidth]{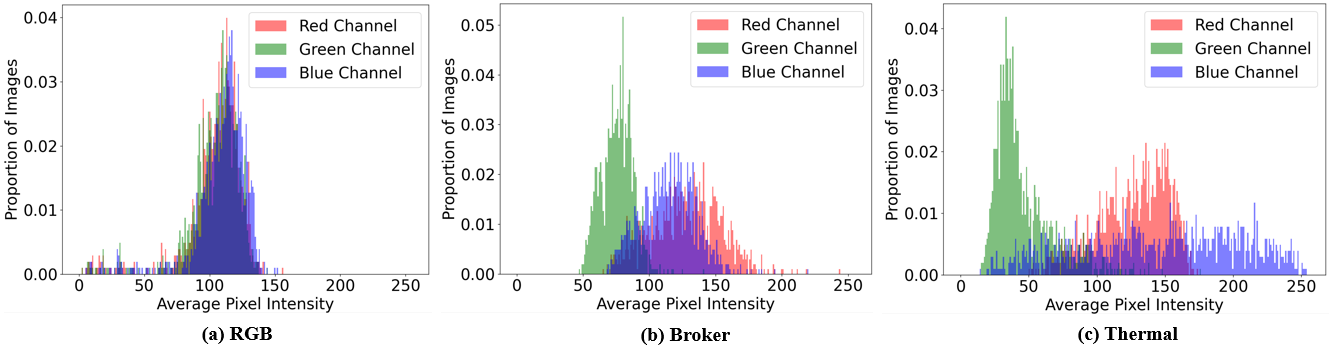}
    \caption{Histograms of the average single-color-channel intensity of all pixels inside an image over the training set of RGBT-CC. They are computed from visible, broker, and thermal images, respectively.}
    \label{fig:wave}
\end{figure}

Moreover, \cref{fig:wave} suggests that BMG finally learns a proper auxiliary modality with an intermediate look between visible and thermal images. With BMG, cross-modal learning becomes much easier as it avoids the whole counting model to deal with the gap between visible and infrared modalities independently.

\begin{table}[tb]
\scriptsize
  \caption{Comparison of different intermediate modality generating methods.}
  \label{tab:5}
  \centering
  \begin{tabular}{c|ccccc|cc}
    \toprule
    Method & GAME(0) & GAME(1) & GAME(2) & GAME(3) & RMSE& Time cost & Params\\
    \hline
    DDFM\cite{zhao2023ddfm}& 11.18 & 14.84 & 19.10 & 25.57 & 18.32&59s&552.81M\\
    BMG&\bf{10.19}& {\bf13.61} & {\bf17.65} & {\bf23.64} & {\bf17.32}&\bf{0.022s}&\bf{3.89M} \\
    \bottomrule
  \end{tabular}
\end{table}

\subsection{Ablation Studies}

We perform a comprehensive analysis and ablation study on RGBT-CC using the proposed model, as shown in \cref{tab:4}.
We here investigate the effectiveness of the structure of BMG. 
From \cref{tab:4}, we notice that compared with using only original modalities, training a lightweight network by distilling DDFM to generate a broker modality enhances the capability of the counting model (\eg GAME(0) is 10.49 and RMSE is 18.03). The performance is clearly improved when we further introduce cross-modal attention (\eg GAME(0) is 10.19 and RMSE is 17.32).
Moreover, as shown in \cref{fig:f}, our BMG alleviates the ghosting effect in the broker modality greatly, especially when the space misalignment is significant (such as in images in the last three lines).
All these results above suggest that BMG enhances the fusion capability of the counting model. With BMG, the whole model better mitigates the detrimental effects of modal misalignment during the fine-tuning stage.

\begin{table}[tb]
\scriptsize
  \caption{The impact of different components of the model on RGBT-CC dataset.}
  \label{tab:4}
  \centering
  \begin{tabular}{c|c|ccccc}
    \toprule
    CMC \& MFD&Cross-modal attention& GAME(0) & GAME(1) & GAME(2) & GAME(3) & RMSE\\
    \hline
    $\times$ & $\times$ & 14.94 & 18.29 & 22.10 & 29.43 & 28.74\\
    \checkmark & $\times$ & 10.49 & 14.11 & 18.28 & 24.37 & 18.03\\   
    \checkmark & \checkmark &{\bf 10.19} & {\bf13.61} & {\bf17.65} & {\bf23.64} & {\bf17.32}\\
  \bottomrule
  \end{tabular}
\end{table}

\section{Conclusion}
\XP{This paper presents a novel approach to multi-modal crowd counting method. By framing the dual-modal visual-thermal crowd counting task as a triple-modal learning problem and introducing a broker modality, we effectively bridge the features of the two original modalities for coherent multi-modal learning. Our proposed broker modality generator, characterized by a non-diffusion, lightweight structure, retains the strong fusion capabilities of denoising diffusion fusion models. Extensive experimental results show that with minor extra parameters, our method outperforms other competitors and achieves promising results. Additionally, we offer a thorough analysis of the ghosting effect inherently in image fusion and justify the effectiveness of the proposed distillation-then-finetuning training strategy for alleviating it.}

\section*{Acknolwedgement}
This work was funded in part by the National Natural Science Foundation of China (62076195, 62376070) and in part by the Fundamental Research Funds for the Central Universities (AUGA5710011522).


\bibliographystyle{splncs04}
\bibliography{egbib}

\begin{thebibliography}{10}
\providecommand{\url}[1]{\texttt{#1}}
\providecommand{\urlprefix}{URL }
\providecommand{\doi}[1]{https://doi.org/#1}

\bibitem{alehdaghi2023adaptive}
Alehdaghi, M., Josi, A., Shamsolmoali, P., Cruz, R.M., Granger, E.: Adaptive generation of privileged intermediate information for visible-infrared person re-identification. arXiv preprint arXiv:2307.03240  (2023)

\bibitem{chen2021topological}
Chen, K., Chen, J.K., Chuang, J., V{\'a}zquez, M., Savarese, S.: Topological planning with transformers for vision-and-language navigation. In: Proceedings of the IEEE/CVF Conference on Computer Vision and Pattern Recognition. pp. 11276--11286 (2021)

\bibitem{fan2020bbs}
Fan, D.P., Zhai, Y., Borji, A., Yang, J., Shao, L.: Bbs-net: Rgb-d salient object detection with a bifurcated backbone strategy network. In: European conference on computer vision. pp. 275--292. Springer (2020)

\bibitem{gao2017image}
Gao, J., Cai, X.f.: Image matching method based on multi-scale corner detection. In: 2017 13th International Conference on Computational Intelligence and Security (CIS). pp. 125--129. IEEE (2017)

\bibitem{guerrero2015extremely}
Guerrero-G{\'o}mez-Olmedo, R., Torre-Jim{\'e}nez, B., L{\'o}pez-Sastre, R., Maldonado-Basc{\'o}n, S., Onoro-Rubio, D.: Extremely overlapping vehicle counting. In: Pattern Recognition and Image Analysis: 7th Iberian Conference, IbPRIA 2015, Santiago de Compostela, Spain, June 17-19, 2015, Proceedings 7. pp. 423--431. Springer (2015)

\bibitem{guo2024consistency}
Guo, Q., Yuan, P., Huang, X., Ye, Y.: Consistency-constrained rgb-t crowd counting via mutual information maximization. Complex \& Intelligent Systems pp. 1--22 (2024)

\bibitem{huang2022reconet}
Huang, Z., Liu, J., Fan, X., Liu, R., Zhong, W., Luo, Z.: Reconet: Recurrent correction network for fast and efficient multi-modality image fusion. In: European Conference on Computer Vision. pp. 539--555. Springer (2022)

\bibitem{idrees2018composition}
Idrees, H., Tayyab, M., Athrey, K., Zhang, D., Al-Maadeed, S., Rajpoot, N., Shah, M.: Composition loss for counting, density map estimation and localization in dense crowds. In: Proceedings of the European conference on computer vision (ECCV). pp. 532--546 (2018)

\bibitem{jiang2020contour}
Jiang, Q., Liu, Y., Yan, Y., Deng, J., Fang, J., Li, Z., Jiang, X.: A contour angle orientation for power equipment infrared and visible image registration. IEEE Transactions on Power Delivery  \textbf{36}(4),  2559--2569 (2020)

\bibitem{kong2024cross}
Kong, W., Liu, J., Hong, Y., Li, H., Shen, J.: Cross-modal collaborative feature representation via transformer-based multimodal mixers for rgb-t crowd counting. Expert Systems with Applications p. 124483 (2024)

\bibitem{korhonen2012peak}
Korhonen, J., You, J.: Peak signal-to-noise ratio revisited: Is simple beautiful? In: 2012 Fourth International Workshop on Quality of Multimedia Experience. pp. 37--38. IEEE (2012)

\bibitem{li2020infrared}
Li, D., Wei, X., Hong, X., Gong, Y.: Infrared-visible cross-modal person re-identification with an x modality. In: Proceedings of the AAAI conference on artificial intelligence. vol.~34, pp. 4610--4617 (2020)

\bibitem{li2022learning}
Li, H., Zhang, S., Kong, W.: Learning the cross-modal discriminative feature representation for rgb-t crowd counting. Knowledge-Based Systems  \textbf{257},  109944 (2022)

\bibitem{li2022rgb}
Li, H., Zhang, S., Kong, W.: Rgb-d crowd counting with cross-modal cycle-attention fusion and fine-coarse supervision. IEEE Transactions on Industrial Informatics  \textbf{19}(1),  306--316 (2022)

\bibitem{li2018dilated}
Li, Y.C.: Dilated convolutional neural networks for understanding the highly congested scenes/y. li, x. zhang, d. chen. In: Proceedings of the IEEE conference on computer vision and pattern recognition.--IEEE. pp. 1091--1100 (2018)

\bibitem{li2020comparison}
Li, Y., Wang, H., Luo, Y.: A comparison of pre-trained vision-and-language models for multimodal representation learning across medical images and reports. In: 2020 IEEE international conference on bioinformatics and biomedicine (BIBM). pp. 1999--2004. IEEE (2020)

\bibitem{lian2021locating}
Lian, D., Chen, X., Li, J., Luo, W., Gao, S.: Locating and counting heads in crowds with a depth prior. IEEE Transactions on Pattern Analysis and Machine Intelligence  \textbf{44}(12),  9056--9072 (2021)

\bibitem{lian2019density}
Lian, D., Li, J., Zheng, J., Luo, W., Gao, S.: Density map regression guided detection network for rgb-d crowd counting and localization. In: Proceedings of the IEEE/CVF Conference on Computer Vision and Pattern Recognition. pp. 1821--1830 (2019)

\bibitem{lin2021direct}
Lin, H., Hong, X., Ma, Z., Wei, X., Qiu, Y., Wang, Y., Gong, Y.: Direct measure matching for crowd counting. the Thirtieth International Joint Conference on Artificial Intelligence  (2021)

\bibitem{lin2024gramformer}
Lin, H., Ma, Z., Hong, X., Shangguan, Q., Meng, D.: Gramformer: Learning crowd counting via graph-modulated transformer. In: Proceedings of the AAAI Conference on Artificial Intelligence. vol.~38, pp. 3395--3403 (2024)

\bibitem{lin2022semi}
Lin, H., Ma, Z., Hong, X., Wang, Y., Su, Z.: Semi-supervised crowd counting via density agency. In: Proceedings of the 30th ACM International Conference on Multimedia. pp. 1416--1426 (2022)

\bibitem{lin2022boosting}
Lin, H., Ma, Z., Ji, R., Wang, Y., Hong, X.: Boosting crowd counting via multifaceted attention. In: Proceedings of the IEEE/CVF Conference on Computer Vision and Pattern Recognition. pp. 19628--19637 (2022)

\bibitem{liu2023point}
Liu, C., Lu, H., Cao, Z., Liu, T.: Point-query quadtree for crowd counting, localization, and more. In: Proceedings of the IEEE/CVF International Conference on Computer Vision. pp. 1676--1685 (2023)

\bibitem{liu2018decidenet}
Liu, J., Gao, C., Meng, D., Hauptmann, A.G.: Decidenet: Counting varying density crowds through attention guided detection and density estimation. In: Proceedings of the IEEE conference on computer vision and pattern recognition. pp. 5197--5206 (2018)

\bibitem{liu2021cross}
Liu, L., Chen, J., Wu, H., Li, G., Li, C., Lin, L.: Cross-modal collaborative representation learning and a large-scale rgbt benchmark for crowd counting. In: Proceedings of the IEEE/CVF conference on computer vision and pattern recognition. pp. 4823--4833 (2021)

\bibitem{Liu_2019_ICCV}
Liu, L., Qiu, Z., Li, G., Liu, S., Ouyang, W., Lin, L.: Crowd counting with deep structured scale integration network. In: Proceedings of the IEEE/CVF International Conference on Computer Vision (ICCV) (October 2019)

\bibitem{liu2018crowd}
Liu, L., Wang, H., Li, G., Ouyang, W., Lin, L.: Crowd counting using deep recurrent spatial-aware network. arXiv preprint arXiv:1807.00601  (2018)

\bibitem{liu2020semi}
Liu, Y., Liu, L., Wang, P., Zhang, P., Lei, Y.: Semi-supervised crowd counting via self-training on surrogate tasks. In: Computer Vision--ECCV 2020: 16th European Conference, Glasgow, UK, August 23--28, 2020, Proceedings, Part XV 16. pp. 242--259. Springer (2020)

\bibitem{liu2023ccanet}
Liu, Y., Cao, G., Shi, B., Hu, Y.: Ccanet: A collaborative cross-modal attention network for rgb-d crowd counting. IEEE Transactions on Multimedia  (2023)

\bibitem{liu2023rgb}
Liu, Z., Wu, W., Tan, Y., Zhang, G.: Rgb-t multi-modal crowd counting based on transformer. The 33rd British Machine Vision Conference 2022  (2022)

\bibitem{ma2015robust}
Ma, J., Zhou, H., Zhao, J., Gao, Y., Jiang, J., Tian, J.: Robust feature matching for remote sensing image registration via locally linear transforming. IEEE Transactions on Geoscience and Remote Sensing  \textbf{53}(12),  6469--6481 (2015)

\bibitem{ma2019bayesian}
Ma, Z., Wei, X., Hong, X., Gong, Y.: Bayesian loss for crowd count estimation with point supervision. In: Proceedings of the IEEE/CVF international conference on computer vision. pp. 6142--6151 (2019)

\bibitem{ma2020learning}
Ma, Z., Wei, X., Hong, X., Gong, Y.: Learning scales from points: A scale-aware probabilistic model for crowd counting. In: Proceedings of the 28th ACM International Conference on Multimedia. pp. 220--228 (2020)

\bibitem{ma2021learning}
Ma, Z., Wei, X., Hong, X., Lin, H., Qiu, Y., Gong, Y.: Learning to count via unbalanced optimal transport. In: Proceedings of the AAAI Conference on Artificial Intelligence. vol.~35, pp. 2319--2327 (2021)

\bibitem{mo2022attention}
Mo, H., Ren, W., Zhang, X., Yan, F., Zhou, Z., Cao, X., Wu, W.: Attention-guided collaborative counting. IEEE Transactions on Image Processing  \textbf{31},  6306--6319 (2022)

\bibitem{mu2024visual}
Mu, B., Shao, F., Xie, Z., Chen, H., Jiang, Q., Ho, Y.S.: Visual prompt multi-branch fusion network for rgb-thermal crowd counting. IEEE Internet of Things Journal  (2024)

\bibitem{pan2024graph}
Pan, Y., Zhou, W., Fang, M., Qiang, F.: Graph enhancement and transformer aggregation network for rgb-thermal crowd counting. IEEE Geoscience and Remote Sensing Letters  (2024)

\bibitem{pan2023cginet}
Pan, Y., Zhou, W., Qian, X., Mao, S., Yang, R., Yu, L.: Cginet: Cross-modality grade interaction network for rgb-t crowd counting. Engineering Applications of Artificial Intelligence  \textbf{126},  106885 (2023)

\bibitem{pang2020hierarchical}
Pang, Y., Zhang, L., Zhao, X., Lu, H.: Hierarchical dynamic filtering network for rgb-d salient object detection. In: Computer Vision--ECCV 2020: 16th European Conference, Glasgow, UK, August 23--28, 2020, Proceedings, Part XXV 16. pp. 235--252. Springer (2020)

\bibitem{peng2020rgb}
Peng, T., Li, Q., Zhu, P.: Rgb-t crowd counting from drone: A benchmark and mmccn network. In: Proceedings of the Asian conference on computer vision (2020)

\bibitem{ren2021learning}
Ren, S., Du, Y., Lv, J., Han, G., He, S.: Learning from the master: Distilling cross-modal advanced knowledge for lip reading. In: Proceedings of the IEEE/CVF Conference on Computer Vision and Pattern Recognition. pp. 13325--13333 (2021)

\bibitem{10.1007/978-3-319-24574-4_28}
Ronneberger, O., Fischer, P., Brox, T.: U-net: Convolutional networks for biomedical image segmentation. In: Navab, N., Hornegger, J., Wells, W.M., Frangi, A.F. (eds.) Medical Image Computing and Computer-Assisted Intervention -- MICCAI 2015. pp. 234--241. Springer International Publishing, Cham (2015)

\bibitem{rublee2011orb}
Rublee, E., Rabaud, V., Konolige, K., Bradski, G.: Orb: An efficient alternative to sift or surf. In: 2011 International conference on computer vision. pp. 2564--2571. Ieee (2011)

\bibitem{sam2020locate}
Sam, D.B., Peri, S.V., Sundararaman, M.N., Kamath, A., Babu, R.V.: Locate, size, and count: accurately resolving people in dense crowds via detection. IEEE transactions on pattern analysis and machine intelligence  \textbf{43}(8),  2739--2751 (2020)

\bibitem{sindagi2019multi}
Sindagi, V.A., Patel, V.M.: Multi-level bottom-top and top-bottom feature fusion for crowd counting. In: Proceedings of the IEEE/CVF international conference on computer vision. pp. 1002--1012 (2019)

\bibitem{tang2022tafnet}
Tang, H., Wang, Y., Chau, L.P.: Tafnet: A three-stream adaptive fusion network for rgb-t crowd counting. In: 2022 IEEE International Symposium on Circuits and Systems (ISCAS). pp. 3299--3303. IEEE (2022)

\bibitem{wang2021self}
Wang, Y., Hou, J., Hou, X., Chau, L.P.: A self-training approach for point-supervised object detection and counting in crowds. IEEE Transactions on Image Processing  \textbf{30},  2876--2887 (2021)

\bibitem{wang2019learning}
Wang, Z., Wang, Z., Zheng, Y., Chuang, Y.Y., Satoh, S.: Learning to reduce dual-level discrepancy for infrared-visible person re-identification. In: Proceedings of the IEEE/CVF conference on computer vision and pattern recognition. pp. 618--626 (2019)

\bibitem{wang2004image}
Wang, Z., Bovik, A.C., Sheikh, H.R., Simoncelli, E.P.: Image quality assessment: from error visibility to structural similarity. IEEE transactions on image processing  \textbf{13}(4),  600--612 (2004)

\bibitem{wei2020co}
Wei, X., Li, D., Hong, X., Ke, W., Gong, Y.: Co-attentive lifting for infrared-visible person re-identification. In: Proceedings of the 28th ACM international conference on multimedia. pp. 1028--1037 (2020)

\bibitem{wu2022multimodal}
Wu, Z., Liu, L., Zhang, Y., Mao, M., Lin, L., Li, G.: Multimodal crowd counting with mutual attention transformers. In: 2022 IEEE International Conference on Multimedia and Expo (ICME). pp.~1--6. IEEE (2022)

\bibitem{xie2023cross}
Xie, Z., Shao, F., Chen, G., Chen, H., Jiang, Q., Meng, X., Ho, Y.S.: Cross-modality double bidirectional interaction and fusion network for rgb-t salient object detection. IEEE Transactions on Circuits and Systems for Video Technology  \textbf{33}(8),  4149--4163 (2023)

\bibitem{xie2024bgdfnet}
Xie, Z., Shao, F., Mu, B., Chen, H., Jiang, Q., Lu, C., Ho, Y.S.: Bgdfnet: Bidirectional gated and dynamic fusion network for rgb-t crowd counting in smart city system. IEEE Transactions on Instrumentation and Measurement  (2024)

\bibitem{xu2023murf}
Xu, H., Yuan, J., Ma, J.: Murf: Mutually reinforcing multi-modal image registration and fusion. IEEE Transactions on Pattern Analysis and Machine Intelligence  (2023)

\bibitem{yang2024cagnet}
Yang, X., Zhou, W., Yan, W., Qian, X.: Cagnet: Coordinated attention guidance network for rgb-t crowd counting. Expert Systems with Applications  \textbf{243},  122753 (2024)

\bibitem{yang2020reverse}
Yang, Y., Li, G., Wu, Z., Su, L., Huang, Q., Sebe, N.: Reverse perspective network for perspective-aware object counting. In: Proceedings of the IEEE/CVF conference on computer vision and pattern recognition. pp. 4374--4383 (2020)

\bibitem{yu2022commercemm}
Yu, L., Chen, J., Sinha, A., Wang, M., Chen, Y., Berg, T.L., Zhang, N.: Commercemm: Large-scale commerce multimodal representation learning with omni retrieval. In: Proceedings of the 28th ACM SIGKDD Conference on Knowledge Discovery and Data Mining. pp. 4433--4442 (2022)

\bibitem{zhang2021mmccn}
Zhang, B., Du, Y., Zhao, Y., Wan, J., Tong, Z.: I-mmccn: Improved mmccn for rgb-t crowd counting of drone images. In: 2021 7th IEEE International Conference on Network Intelligence and Digital Content (IC-NIDC). pp. 117--121. IEEE (2021)

\bibitem{zhang2020uc}
Zhang, J., Fan, D.P., Dai, Y., Anwar, S., Saleh, F.S., Zhang, T., Barnes, N.: Uc-net: Uncertainty inspired rgb-d saliency detection via conditional variational autoencoders. In: Proceedings of the IEEE/CVF conference on computer vision and pattern recognition. pp. 8582--8591 (2020)

\bibitem{zhang2019wide}
Zhang, Q., Chan, A.B.: Wide-area crowd counting via ground-plane density maps and multi-view fusion cnns. In: Proceedings of the IEEE/CVF Conference on Computer Vision and Pattern Recognition. pp. 8297--8306 (2019)

\bibitem{zhang2022spatio}
Zhang, Y., Choi, S., Hong, S.: Spatio-channel attention blocks for cross-modal crowd counting. In: Proceedings of the Asian Conference on Computer Vision. pp. 90--107 (2022)

\bibitem{zhang2021towards}
Zhang, Y., Yan, Y., Lu, Y., Wang, H.: Towards a unified middle modality learning for visible-infrared person re-identification. In: Proceedings of the 29th ACM International Conference on Multimedia. pp. 788--796 (2021)

\bibitem{zhao2023metafusion}
Zhao, W., Xie, S., Zhao, F., He, Y., Lu, H.: Metafusion: Infrared and visible image fusion via meta-feature embedding from object detection. In: Proceedings of the IEEE/CVF Conference on Computer Vision and Pattern Recognition. pp. 13955--13965 (2023)

\bibitem{zhao2023cddfuse}
Zhao, Z., Bai, H., Zhang, J., Zhang, Y., Xu, S., Lin, Z., Timofte, R., Van~Gool, L.: Cddfuse: Correlation-driven dual-branch feature decomposition for multi-modality image fusion. In: Proceedings of the IEEE/CVF Conference on Computer Vision and Pattern Recognition. pp. 5906--5916 (2023)

\bibitem{zhao2023ddfm}
Zhao, Z., Bai, H., Zhu, Y., Zhang, J., Xu, S., Zhang, Y., Zhang, K., Meng, D., Timofte, R., Van~Gool, L.: Ddfm: denoising diffusion model for multi-modality image fusion. In: Proceedings of the IEEE/CVF International Conference on Computer Vision. pp. 8082--8093 (2023)

\bibitem{zhou2022defnet}
Zhou, W., Pan, Y., Lei, J., Ye, L., Yu, L.: Defnet: Dual-branch enhanced feature fusion network for rgb-t crowd counting. IEEE Transactions on Intelligent Transportation Systems  \textbf{23}(12),  24540--24549 (2022)

\bibitem{zhou2024mjpnet}
Zhou, W., Yang, X., Dong, X., Fang, M., Yan, W., Luo, T.: Mjpnet-s*: Multistyle joint-perception network with knowledge distillation for drone rgb-thermal crowd density estimation in smart cities. IEEE Internet of Things Journal  (2024)

\bibitem{zhou2023mc}
Zhou, W., Yang, X., Lei, J., Yan, W., Yu, L.: $\mathrm{MC^3Net}$: Multimodality cross-guided compensation coordination network for rgb-t crowd counting. IEEE Transactions on Intelligent Transportation Systems  (2023)

\end{thebibliography}

\clearpage 
\appendix

\setcounter{table}{0}
\setcounter{figure}{0}
\setcounter{section}{0}
\setcounter{equation}{0}

\section*{Appendix}
\section{Additional Experiments}

{Here we present two experiments to further demonstrate our proposed broker modality generator (BMG) and the `two-stage' learning scheme. 
First, we evaluate the image registration-based methods for ghost effect alleviation on crowd images and find that these methods perform poorly. This underscores the necessity of designing a new method instead of using traditional alignment methods to mitigate space misalignment between two modalities (as described in footnote 2 in the main paper), which motivates the design of our method. 
Next, we demonstrate the improvement in the quality of the intermediate-modal images, which further highlights the effectiveness of our proposed BMG and the two-stage learning scheme in mitigating the ghosting effect. Further details are elaborated below.}

\subsection{Illustration of the Performance of Image Registration Algorithms on Crowd Images}

The primary goal of image registration is to align and superimpose multiple images depicting the same scene but captured under varying resolutions, perspectives, or fields of view. Among all vision-infrared registration approaches, point feature matching \cite{ma2015robust,jiang2020contour,gao2017image} is the most popular, whose target is to find matching points between vision and infrared images and obtain the transformation parameters such as shifts, rotation, and shear.

Initially, we strive to align visual and thermal images through registration before feeding them to the fusion model, aiming to mitigate the space misalignment and thereby mitigate the ghosting phenomenon in the intermediate modality.
Unfortunately, we have found that despite the satisfactory outcomes achieved by the registration methods mentioned above on natural images, they perform inadequately on crowd image pairs with low capture quality and poor visibility conditions. We will elaborate on the experimental results and detailed rationale in the following section.

\begin{figure}[htb]
    \centering
    \includegraphics[width=0.8\textwidth]{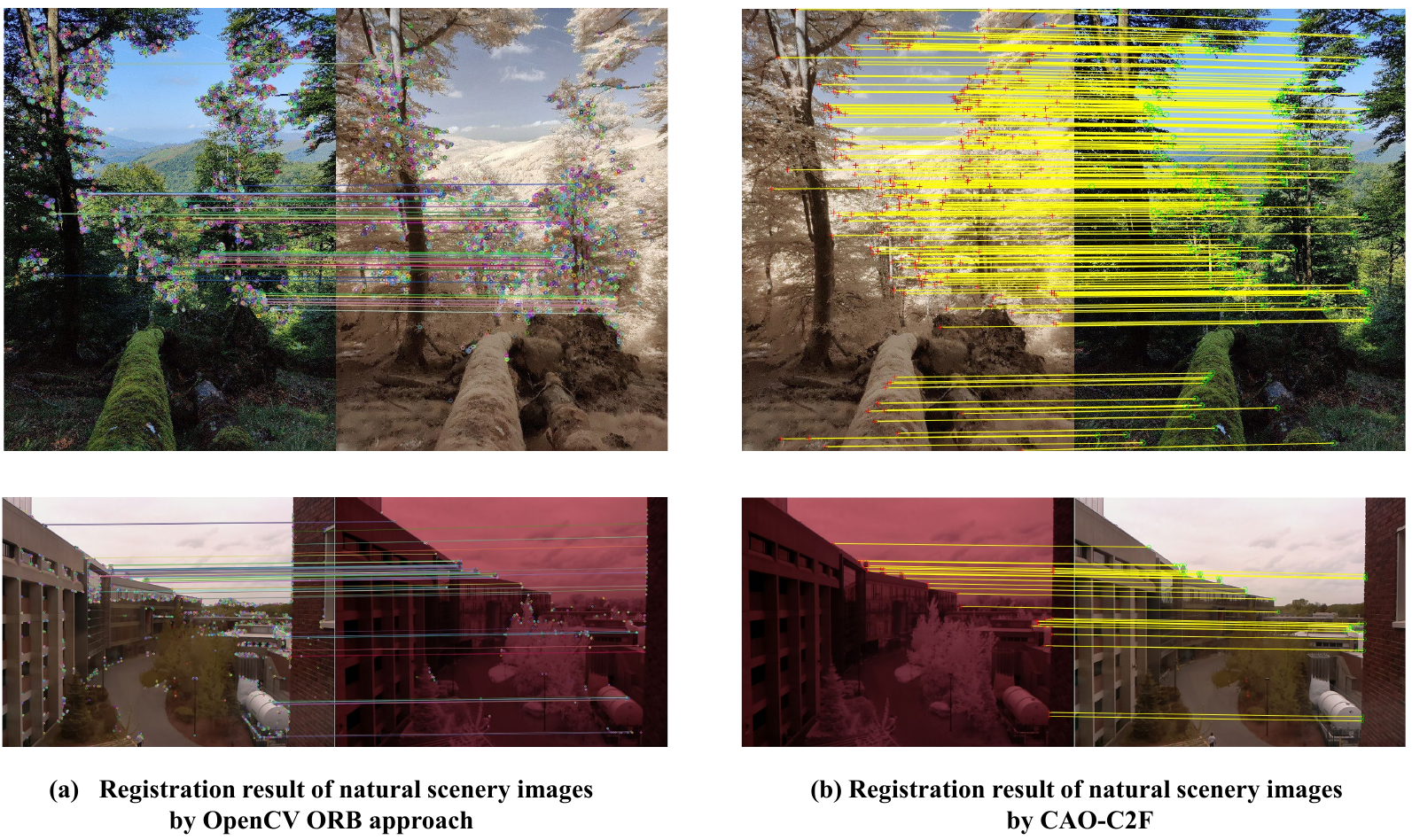}
    \caption{Registration results of natural scenery images by OpenCV ORB approach\cite{rublee2011orb} (a) and CAO-C2F\cite{jiang2020contour} (b). Feature points are marked with dots and matched feature points are connected using line segments. For clarity, we only indicate the pairing points with the top 1\% highest matching confidence points. Registration algorithms match feature points accurately on natural scenery images.}
    \label{fig:r1}
\end{figure}
\begin{figure}[htb]
    \centering
    \includegraphics[width=0.8\textwidth]{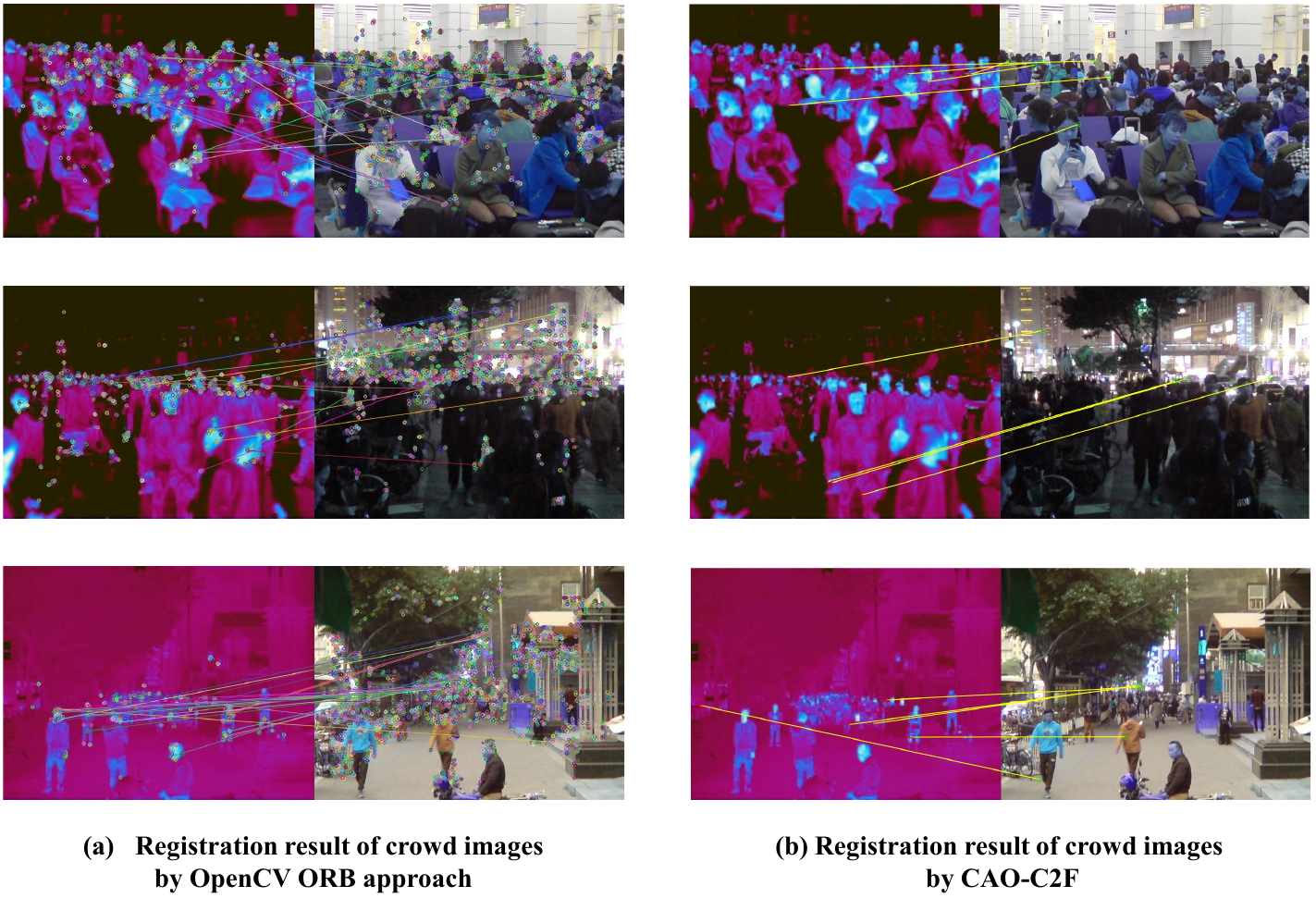}
    \caption{Registration results of crowd images from the testing set of RGBT-CC \cite{liu2021cross} by OpenCV ORB approach \cite{rublee2011orb}(a) and CAO-C2F \cite{jiang2020contour} (b). Feature points are marked with dots and matched feature points are connected using line segments. For clarity, we only indicate the pairing points with the top 1\% highest matching confidence points. Registration algorithms perform inadequately on crowd images.}
    \label{fig:r2}
\end{figure}

\subsubsection{Results.}
We attempt to align visual and thermal images utilizing ORB \cite{rublee2011orb} \footnote{We complete our experiment by an open source script using the ORB approach implemented by OpenCV. The script is available at \url{https://github.com/khufkens/align\_images}.} and CAO-C2F \cite{jiang2020contour}, and the results are shown in \cref{fig:r2}. For comparison, we also conduct experiments on natural scenery images, as shown in \cref{fig:r1}. The results suggest that modern registration approaches perform well on natural scenery images. Most line segments connecting paired points in \cref{fig:r1} are horizontal and parallel, accurately matching the corresponding feature points in different modalities. However, these approaches perform inadequately on crowd images. Most line segments connecting paired points in \cref{fig:r2} are sloping and messy. It struggles to distinguish different entities in the crowd images and correctly match them. 

\subsubsection{Discussion.}
Most restriction approaches are POI methods, that is, alignments are based on Point Of Interest. Their alignment effect is unsatisfactory on the crowd images we use in our study. First, due to low imaging quality such as poor illumination conditions and low resolution, it is difficult to extract enough features from the crowd for matching feature points, causing the alignment output inaccurate. Besides, it is hard to distinguish between different individuals in dense crowds, especially in thermal images. Thus, the restriction methods often match one interest point of a certain person to the same body part of another person because of the feature similarity between different individuals.

In general, existing registration methods are not well-suited for aligning the crowd image pairs. Alleviating the ghosting effect in the fused image in multi-modal crowd counting through visible-infrared image registration is impractical. This presents us with the opportunity to introduce our training scheme to eliminate the ghosting phenomenon.

\subsection{\XP{Comparisons on the Quality of Intermediate-Modal Images}}

\begin{table}[htb]
\scriptsize

  \caption{Comparison of the quality of intermediate-modal images on the testing set of RGBT-CC \cite{liu2021cross}.}
  \label{tab:psnr}
  \centering
  \begin{tabular}{@{}cc|cc|cc@{}}
\toprule
\multicolumn{2}{c|}{Training scheme}&\multicolumn{2}{c|}{BMG Structure}&\multirow{2}{*}{PSNR(dB)$\uparrow$}&\multirow{2}{*}{SSIM$\uparrow$}\\

\cline{1-4}
Distillation& Fine-tuning&UNet&CMA& \\
\cline{1-6}
$\times$&$\times$&-&-&15.36&0.59\\
\checkmark&$\times$&\checkmark&$\times$&15.97&0.61\\
\checkmark&$\times$&\checkmark&\checkmark&15.74&0.60\\
\checkmark&\checkmark&\checkmark&$\times$&23.75&0.83\\
\checkmark&\checkmark&\checkmark&\checkmark&\textbf{26.67}&\textbf{0.88}\\ 
\bottomrule
\end{tabular}
\end{table}

We compare \XP{the intermediate-modal images in terms of two performance metrics:} the Peak Signal-to-Noise Ratio (PSNR) \cite{korhonen2012peak} and Structural Similarity Index Measure (SSIM) \cite{wang2004image} on the testing set of RGBT-CC \cite{liu2021cross} to evaluate the effectiveness of ghosting effect alleviation, as shown in \cref{tab:psnr}. The result demonstrates that the two-stage training scheme facilitates the elimination of the ghosting effect caused by directly fusing two original modalities. First, we can notice that without the fine-tuning stage (in line 2 and 3 of \cref{tab:psnr}), BMG can not acquire the capability to alleviate the ghosting phenomenon.
Next, after the fine-tuning stage, the reconstruction quality of our broker modality significantly improves. Compared with the images generated directly by DDFM\cite{zhao2023ddfm} (in the first line of \cref{tab:psnr}), the ghosting effect in the broker modal is greatly eliminated. 
Moreover, when we further introduce the cross-modal attention module (CMA), the values of both PSNR and SSIM increase. It suggests that with cross-modal attention, our BMG acquires a higher capability to alleviate the ghosting effect in the fine-tuning stage. In general, our proposed two-stage learning scheme shows a great effect on alleviating the ghosting phenomenon caused by directly fusing two original modalities with angular disparity or spatial displacement.

\end{document}